\documentclass[10pt,twocolumn,letterpaper]{article}

\usepackage{iccv}              

\usepackage{pifont}
\usepackage{dsfont}
\usepackage{adjustbox}
\usepackage{multicol}
\usepackage{multirow}
\usepackage{algorithm}
\usepackage{algpseudocode}
\usepackage{makecell}
\usepackage{booktabs}
\usepackage[T1]{fontenc}
\usepackage{colortbl}

\definecolor{iccvblue}{rgb}{0.21,0.49,0.74}
\usepackage[pagebackref,breaklinks,colorlinks,allcolors=iccvblue]{hyperref}

\def\paperID{13933} 
\def\confName{ICCV}
\def\confYear{2025}

\newcommand{\methodname}{\texttt{TruthPrInt}\xspace}
\newcommand{\alignmethodname}{\texttt{ComnHallu}\xspace}

\usepackage{amsmath,amsfonts,bm}

\def\eqref#1{equation~\ref{#1}}

\def\1{\bm{1}}

\def\vc{{\bm{c}}}

\def\vh{{\bm{h}}}

\def\vk{{\bm{k}}}

\def\vo{{\bm{o}}}

\def\vr{{\bm{r}}}
\def\vs{{\bm{s}}}

\def\vx{{\bm{x}}}

\def\vz{{\bm{z}}}

\DeclareMathAlphabet{\mathsfit}{\encodingdefault}{\sfdefault}{m}{sl}
\SetMathAlphabet{\mathsfit}{bold}{\encodingdefault}{\sfdefault}{bx}{n}

\DeclareMathOperator*{\argmin}{arg\,min}

\title{\methodname: Mitigating Large Vision-Language Models Object Hallucination Via Latent Truthful-Guided Pre-Intervention}

\author{Jinhao Duan$^1$\thanks{Equal Contribution}, Fei Kong$^{2\,*}$, Hao Cheng$^3$, James Diffenderfer$^4$, Bhavya Kailkhura$^4$, \\ Lichao Sun$^5$, Xiaofeng Zhu$^2$, Xiaoshuang Shi$^2$, Kaidi Xu$^1$\thanks{Correspondence to Kaidi Xu \texttt{kx46@drexel.edu}} \\
$^1$Drexel University $^2$University of Electronic Science and Technology of China \\ $^3$Hong Kong University of Science and Technology (Guangzhou) $^4$LLNL $^5$Lehigh University 
}

\begin{document}
\maketitle
\begin{abstract}
Object Hallucination (OH) has been acknowledged as one of the major trustworthy challenges in Large Vision-Language Models (LVLMs).
Recent advancements in Large Language Models (LLMs) indicate that internal states, such as hidden states, encode the ``overall truthfulness'' of generated responses. However, it remains under-explored how internal states in LVLMs function and whether they could serve as ``per-token'' hallucination indicators, which is essential for mitigating OH.
In this paper, we first conduct an in-depth exploration of LVLM internal states with OH issues and discover that \ding{202} LVLM internal states are \underline{high-specificity} per-token indicators of hallucination behaviors. Moreover, \ding{203} different LVLMs encode \underline{universal patterns} of hallucinations in common latent subspaces, indicating that there exist ``generic truthful directions'' shared by various LVLMs. 
Based on these discoveries, we propose \underline{Truth}ful-Guided \underline{Pr}e-\underline{Int}ervention (\textbf{\methodname}) that first learns the truthful direction of LVLM decoding and then applies truthful-guided inference-time intervention during LVLM decoding. We further propose \textbf{\alignmethodname} to enhance both cross-LVLM and cross-data hallucination detection transferability by constructing and aligning hallucination latent subspaces.
We evaluate \methodname in extensive experimental settings, including in-domain and out-of-domain scenarios, over popular LVLMs and OH benchmarks. Experimental results indicate that \methodname significantly outperforms state-of-the-art methods. Codes will be available at \url{https://github.com/jinhaoduan/TruthPrInt}.

\end{abstract}    
\section{Introduction}
\label{sec:intro}

As Large Vision-Language Models (LVLMs)~\cite{zhuminigpt,liu2024visual,ye2024mplug} have rapidly advanced in cross-modal content understanding and instruction following, their trustworthiness is threatened by Object Hallucination (OH)~\cite{rohrbach2018object,rawte2023survey}.
Although recent work reveals that Large Language Model (LLM) internal states, such as hidden states, entail richer semantic and contextual information~\cite{du2024haloscope,chen2024inside,zhu2024pollmgraph,azaria2023internal,chen2023beyond,li2024reference} that can reveal the truthfulness (or uncertainty)~\cite{kuhnsemantic,lingenerating,duan2024shifting} of model generations, it remains under-explored \textbf{\textit{(i)}} whether internal states in LVLMs encode information about truthfulness, \textbf{\textit{(ii)}} whether they support ``per-token'' hallucination analysis, and \textbf{\textit{(iii)}} whether this information can be transferred to enhance practical applications, e.g., Out-of-Distribution (OOD) shifting.
Preliminary research on internal states of LVLM relies mainly on statistical aspects of internal states to identify hallucinations, such as self-attention activation patterns~\cite{huang2024opera,gong2024damro} and long-term decay~\cite{xing2024mitigating} in RoPE~\cite{su2024roformer}. However, these approaches do not explicitly link hidden states to hallucination behaviors and tend to be effective only for specific datasets and model architectures.

In this paper, we investigate: \textbf{Are internal states reliable and practical indicators of LVLM per-token hallucination behaviors?}
To answer this, we first create datasets consisting of thousands of internal states, each labeled with hallucination membership, i.e., as truthful or hallucinated. By training models on it for hallucination detection, we observe that \ding{202} LVLM internal states provide \underline{undesirable overall performance} yet they are \underline{high-specificity} indicators: the Likelihood Ratio for Positive Results (LR+) achieves nearly $20$, indicating internal states provide confident detection with extremely low false alarm; \ding{203} There exist latent common hallucination subspaces shared by different LVLMs, in which detectors trained on the projections in this subspace are capable of transferring to OOD domains. This suggests the existence of common ``truthful directions'' shared by various LVLMs.

\begin{figure*}[h]
    \centering
    \includegraphics[width=0.95\linewidth]{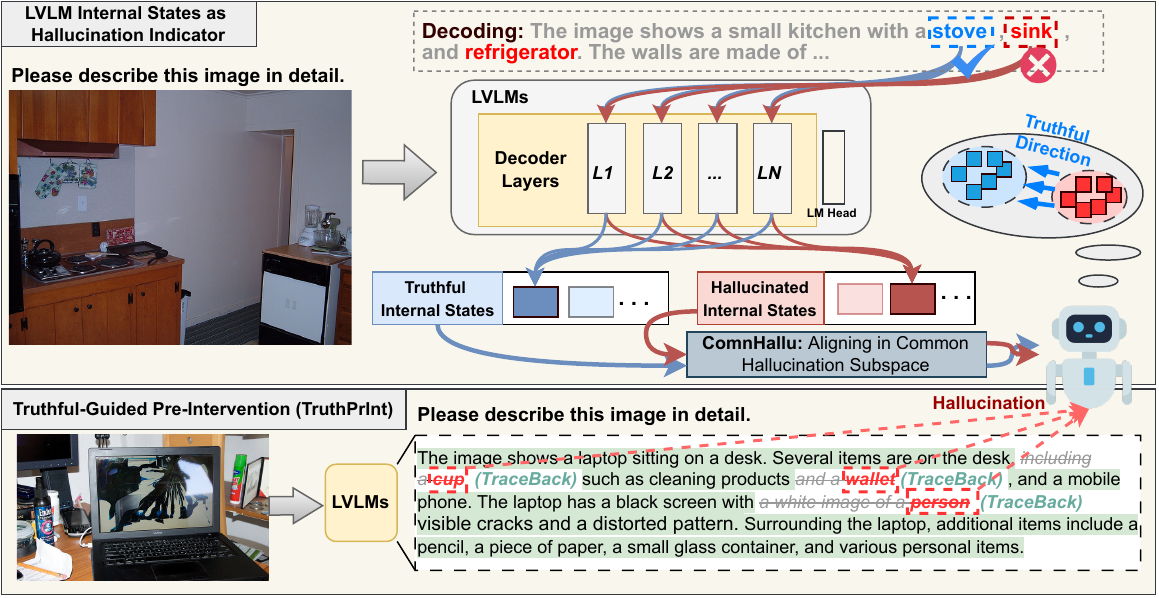}
    \caption{The overall pipeline of \methodname for OH mitigation. \methodname first collects internal states from LVLMs and learns ``truthful direction'' from the latent space. A subspace alignment method \alignmethodname is also proposed to enhance testing-time transferability among various LVLMs and datasets. During decoding, \methodname guides the target VLM towards the truthful direction by rejecting hallucinated tokens and tracing back to ``early starting points'' for pre-intervention.}
    \label{fig:overall}
    \vspace{-4mm}
\end{figure*}

Based on these, we design \textbf{\methodname}, a novel two-stage OH mitigation framework: first locating hallucinated tokens from latent subspace and then performing truthful-guided interventions enabling truthful decoding. We also propose \textbf{\alignmethodname}, a hallucination subspace alignment method, to improve OOD transferability for hallucination detection.
\methodname is evaluated on advanced LVLMs including MiniGPT-4~\cite{zhuminigpt}, Llava-1.5~\cite{liu2024visual}, mPLUG-Owl2~\cite{ye2024mplug}, QWen2VL~\cite{wang2024qwen2}, InternVL-2.5~\cite{chen2024expanding}, over popular OH benchmarks such as CHAIR~\cite{rohrbach2018object}, POPE~\cite{li2023evaluating}, and LLaVA-Bench~\cite{liu2024improved}. Experimental results show that \methodname significantly outperforms competitive baselines and verified on both in-domain and OOD scenarios.
Our contribution can be summarized as the following:
\begin{itemize}
    \item We provide an in-depth exploration of how LVLM internal states related to OH and found that internal states are high-specificity hallucination indicators, encoding universal hallucination patterns from various LVLMs.
    \item We propose a novel two-stage framework \methodname to mitigate OH in LVLMs, and~\alignmethodname, capturing common hallucination features from subspace to enhance cross-LVLM and cross-data transferability.
    \item We conduct comprehensive experiments on popular LVLMs and  OH benchmarks. Experimental results indicate that \methodname significantly outperforms advanced baselines.
\end{itemize}

\section{Related Work}
\label{sec:related_work}

\noindent\textbf{Object Hallucination in LVLMs.}
Object Hallucination (OH)~\cite{rawte2023survey} typically refers to the phenomenon where LVLMs generate nonexistent visual elements, such as objects~\cite{rohrbach2018object}, attributes~\cite{feng2024more}, or events~\cite{zhang2024eventhallusion}, posing a significant challenge to achieving trustworthy performance. A considerable of benchmarks~\cite{wu2024autohallusion,rohrbach2018object,Fu2023MMEAC,li2023evaluating,wu2024unified,chen2024multi,wang2024mitigatingb,villa2023behind,fu2024video} are proposed for OH evaluation, such as CHAIR~\cite{rohrbach2018object}, MME~\cite{Fu2023MMEAC}, and POPE~\cite{li2023evaluating}. To mitigate OH, two lines of research are proposed for OH mitigation: \textit{Contrastive Decoding} (CD)~\cite{o2023contrastive,chenhalc,chuangdola,leng2024mitigating,wan2024contrastive,kim2024code,chen2024alleviating,wang2024mitigating,liu2024paying} and \textit{post-processing}~\cite{yin2023woodpecker,zhouanalyzing,wu2024logical,ouali2024clip}.
CD primarily reduces biases imposed in LVLMs by contrasting generated responses from various decoding strategies, including distinct visual regions~\cite{wan2024contrastive,chenhalc,leng2024mitigating,liu2024paying}, self-contrastive~\cite{chuangdola,kim2024code,wang2024mitigating}, and contrasting with preference models~\cite{chen2024alleviating}. 
CD approaches for OH mitigation are sensitive to specific contrasting objects and often rely on a narrow set of biases, overlooking the complex factors that contribute to LVLM hallucinations.
Post-processing methods~\cite{yin2023woodpecker,zhouanalyzing,deng2024seeing} usually apply iterative visual prompting and continuous editing of the generated response. These methods may bring considerable computational overhead and are often designed for specific tasks.

\noindent\textbf{Internal Representations in Language Models.} Internal representations typically refer to intermediate model outputs, such as self-attention maps and hidden states~\cite{vaswani2017attention}. These representations have been widely used to study language model behaviors, including knowledge (or neuron) editing~\cite{meng2022locating,meng2022memit}, enhancing inference-time reasoning~\cite{li2024inference}, and enabling interpretability~\cite{zhao2024explainability}. In terms of hallucination modeling, recent research indicates that internal representations—like hidden states~\cite{chen2024inside} and attention head activations~\cite{li2024inference}—contain more ``truthfulness'' information than generated textual responses. Building on this insight, substantial work~\cite{du2024haloscope,chen2024inside,zhu2024pollmgraph,azaria2023internal,chen2023beyond,li2024reference} has focused on language model uncertainty quantification (UQ)~\cite{kuhnsemantic,lingenerating,duan2024shifting}, by either measuring the semantic consistency~\cite{chen2024inside} of hidden states or training detectors explicitly designed to identify overall hallucination behaviors~\cite{du2024haloscope,zhu2024pollmgraph,li2024reference}. 

However, UQ focuses on the ``overall truthfulness'' of generated responses. How internal states of LVLMs function within OH remains unclear.
Current studies primarily depend on simple statistical metrics, such as self-attention activation patterns~\cite{huang2024opera,gong2024damro} and long-term decay~\cite{xing2024mitigating} in RoPE, to detect hallucinations. These methods are typically effective only for certain datasets and specific model architectures. Nullu~\cite{yang2024nullu} identifies the hallucination subspace within the latent space and edits LVLMs away from it to achieve truthful decoding. However, this process may considerably impact LLM benign behaviors, as previously highlighted in knowledge-editing research~\cite{liunveiling}.
Differently, our work directly models LVLM hallucination behaviors using internal states with per-token annotations and additionally offers guidance for decoding to reduce OH.

\section{Modeling Transferable LVLM Hallucination Features in Common Latent Subspace}\label{sec:detection}

In this section, we demonstrate that internal states are reliable indicators of LVLM per-token hallucination behaviors. Additionally, we identify the existence of latent subspace that contains transferable hallucination features, enabling the hallucination detector to generalize across different datasets and models. 

\subsection{Crafting Per-Token Hallucination Detector}\label{sec:classifier}
\noindent\textbf{Internal States Collection.} To enable per-token hallucination detection, we first craft LVLM internal states and the corresponding hallucination labels. We prompt LVLM to describe images from the CC-Sbu-Align~\cite{zhuminigpt} dataset, which consists of 3,439 detailed image-description pairs from Conceptual Captions~\cite{sharma2018conceptual,changpinyo2021conceptual} and SBU~\cite{ordonez2011im2text}.
Specifically, for given LVLM $\mathcal{M}$ parameterized by $\bm\theta$, image $\vx$, and prompt $\vs$ for description, the $i$-th generated token is denoted by $z_i = p_{\bm\theta}(\cdot|\vx, \vz_{<i}, \vs)$ where $z_{<i}$ refers to the previously generated $i-1$ tokens. The hidden state of token $z_i$ is denoted by $\vh_{z_i}^{l} = \mathcal{M}^{l}(\vx, \vs, \vz_{<i+1};\bm\theta)\,\,(1\leq i \leq n, \,\, \vh^{l}_{z_i} \in \mathbb{R}^{d})$ where $n$ is the length of the generated tokens and $d$ is the hidden state dimension, e.g., $d=4,096$ in MiniGPT-4.
A token $z_i$ is identified as an \textit{object token} $z^{o}_{i}$ if $z_i$ completes a noun. Then, for each object token $z^{o}_{i}$, we collect the \underline{hidden states of its previous token}, i.e., hidden states $\vh^{l}_{z_{i-1}}$ whose ``next-token'' prediction resulting $z^{o}_i$, as the target internal states.
The reason we collect ``previous hidden states'' rather than current object hidden states is two-fold: 
\ding{202} This one-step-ahead approach allows the detector to provide early warnings of potential hallucinations and enable it to learn general patterns where hallucinations may occur rather than identifying specific hallucinated tokens; \ding{203} Enabling conveniently hidden state intervention for truthful next-token decoding (~\cref{sec:intervention}). Please refer to~\cref{appendix:collect_hs} for more discussion. Next, each hidden state $\vh^l_i$ is equipped with a membership $y_i$: hallucinated if the corresponding object does not appear in the image's reference description, i.e., $y_i=1$, or truthful, i.e., $y_i=0$. Eventually, we collected balanced internal states datasets from MiniGPT-4, Llava-1.5, and mPLUG-Owl2, e.g., 2,716 hallucinated and truthful internal states, respectively, from MiniGPT-4.

\begin{figure}[t]
    \centering
    \includegraphics[width=\linewidth]{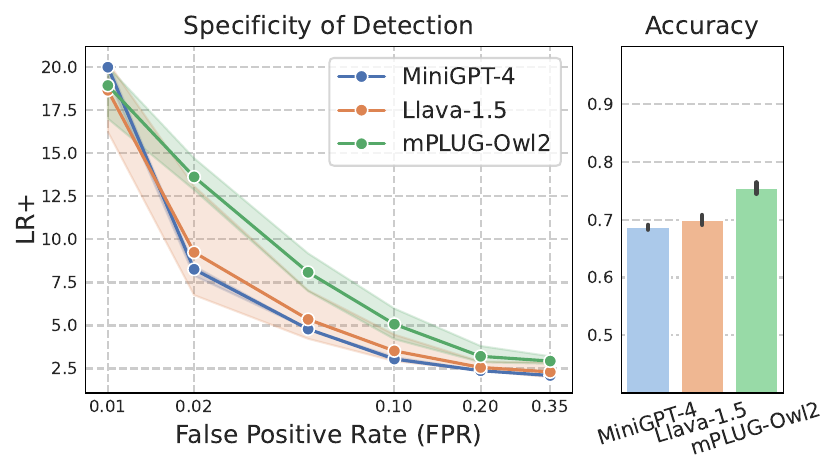}
    \vspace{-7mm}
    \caption{The performance of the designed hallucination detector across various LVLMs. Although internal states offer limited discriminative features for overall accuracy, they achieve high-specificity detections with low false alarm rates.}
    \label{fig:highspecificity}
    \vspace{-3mm}
\end{figure}

\noindent\textbf{Hallucination Detection.} Formally, we denote by $\mathcal{H}=\{\vh^{l}_{i} \in \mathbb{R}^{d}: y_{i} = 1\}$ the set of hallucinated internal states and $\mathcal{T}=\{\vh^{l}_{i} \in \mathbb{R}^{d}: y_{i} = 0\}$ the set of truthful internal states. The hallucination detection~\cite{du2024haloscope} is then formulated as optimizing model $\mathcal{G}_{\bm\theta}$ to miminizing risk
\begin{equation}
\begin{aligned}
\small
\mathcal{R}_{\mathcal{H}, \mathcal{T}} & = \mathcal{R}^{+}_{\mathcal{H}}(\mathcal{G}) + \mathcal{R}^{-}_{\mathcal{T}}(\mathcal{G})\\
& =\mathbb{E}_{\vh \sim \mathcal{H}}\mathds{1}\{\mathcal{G}(\vh) \leq 0\} + \mathbb{E}_{\vh \sim \mathcal{T}}\mathds{1}\{\mathcal{G}(\vh) > 0\}
\end{aligned}
\end{equation}
The hallucination membership of a testing sample $\vh$ is given by $\mathbf{H}(\vh) = \mathds{1}\left[ \mathcal{G}(\vh) \geq \tau \right]$, where $\tau$ is the threshold. In our implementation, $\mathcal{G}$ is a 3-layer MLP, taking the middle layer hidden state as input, i.e., $l=16$, trained with Binary Cross Entropy (BCE) loss. We use 80\% of collected internal states for training and 20\% for validation. Please refer to~\cref{appendix:classifier_training} for a detailed training protocol.

\begin{figure*}[ht]
        \centering
        \includegraphics[width=0.95\linewidth]{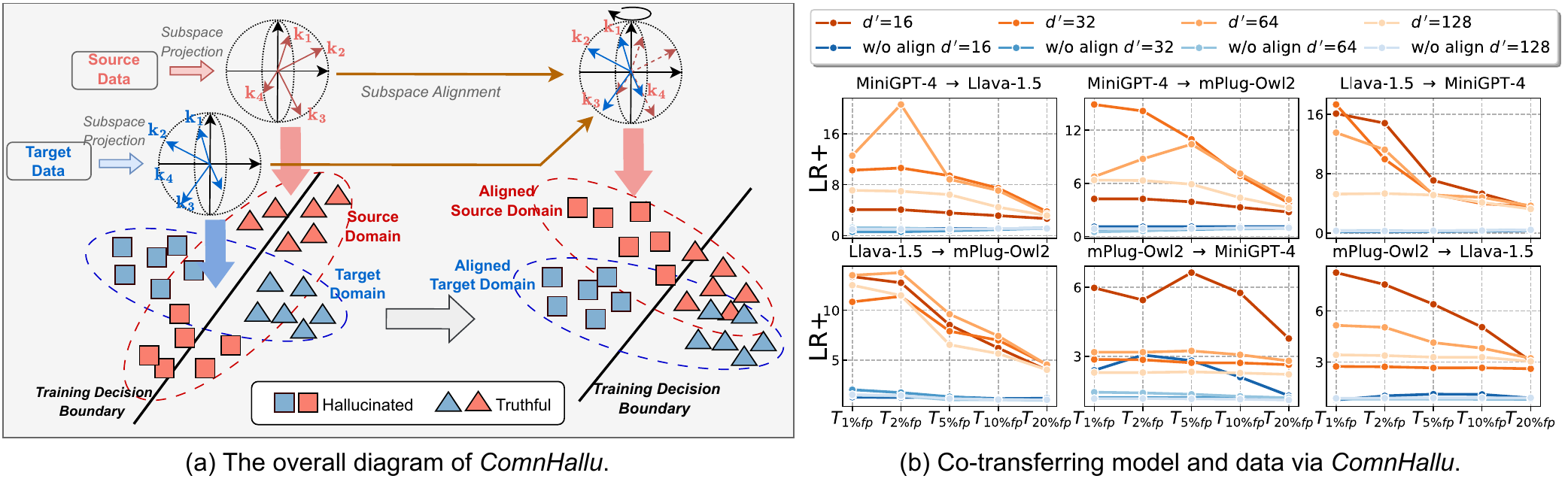}
        \caption{\alignmethodname (a) identifies common latent subspaces shared by both target (training) domain and source (testing) domain, capturing hallucination features, which (b) maintains internal states to be high-specificity when transferring both data domain and models. $T_{\alpha\, \text{fp}}$ means the threshold resulting $\text{FPR}=\alpha$ in the CC-Sbu-Align validation set.}\label{fig:comnhallu}
        \vspace{-5mm}
\end{figure*}

\subsection{Mitigate OH Needs High-Specificity Indicator}
In OH, object tokens only take an extremely small portion of generated tokens, e.g., $\sim$5.6\% tokens are object tokens in MiniGPT-4 captions, and $\sim$10\% among them are hallucinated. Thus, it is essential to make the hallucination detector \underline{high-specificity}, i.e. low False Positive Rate (FPR) while maintaining a certain True Positive Rate (TPR) to reduce false alarm examples, which is different from ``overall truthfulness (or uncertainty)'' quantification in LLMs. To evaluate this, we employ \textit{Likelihood Ratio for Positive Results (LR+)} as the metric where $\textit{LR+}=\textit{TPR} / \textit{FPR}$. Results are summarized in~\cref{fig:highspecificity}. Accuracy is calculated by classifying the top 50\% of predictions as hallucinated and the remaining 50\% as truthful. It is shown that internal states offer limited discriminative features for overall accuracy (error rate $> 20\%$ greater than the portion of hallucinated tokens). However, they achieve near 20 LR+ at FPR=0.01, meaning that our crafted internal states are high-specificity indicators of hallucination.

\subsection{Transferable Hallucination Detection via Subspace Alignment}\label{sec:comnhallu_transfer}
It is crucial that the hallucination detector remains robust under \underline{domain shifting}, i.e., the training (or \textit{target}) domain of the hallucination detector is different from the data and models in the testing (or \textit{source}) domain. However, as shown in~\cref{fig:comnhallu} (b), original internal states (blue curves) show poor transferability when transferring training domains to testing domains.

Recent research shows that LLMs encode similar semantics across various backbone models, e.g., invariant relative representation~\cite{moschellarelative,huang2024enabling} and occasionally exhibit similar types of flaws~\cite{puchert2023llmmaps}, such as LLMs comparing 9.11 and 9.9~\cite{xie2024order}. This indicates that different LVLMs may share common OH features.
Inspired by this, we design \textbf{\alignmethodname}, a straightforward unsupervised domain adaptation method that identifies a common latent subspace containing shared hallucination features between the source and target domains.

\alignmethodname first identifies base vectors separately from the training and testing domains, then projects all hidden states into the respective subspaces defined by these base vectors. Next, a linear transformation is applied to align the testing domain's base vectors with those from the training domain. This alignment ensures that hidden states from the testing domain can be represented using bases that are close to the training domain bases, thus achieving distributional alignment between the projected hidden states of both domains. The overall framework is presented in~\cref{fig:comnhallu} (a).

Concretely, given $N$ internal states $\{\vh_i\}_i^\textit{N}$ (layer index $l$ is omitted) sampled from source domain $\mathcal{S} \subseteq \mathbb{R}^d$ and $M$ internal states $\{\vh_i\}_i^\textit{M}$ from target domain $\mathcal{D} \subseteq \mathbb{R}^d$, the task is to identify a subspace $\mathcal{C} \subseteq \mathbb{R}^{d^\prime} \,\, (d^\prime < d)$ such that \textit{\textbf{(i)}} projections of internal states from both domains onto $\mathcal{C}$ should retain hallucination-related features; \textit{\textbf{(ii)}} projections of the source and target internal states should follow a similar distribution within $\mathcal{C}$, i.e., distribution alignment.

We stack source internal states into feature matrices: $\mathbf{S} \in \mathbb{R}^{N\times d}$, and pre-process them to be 0-centered and normalize each $\vh_i$ by its Frobenius norm: $\tilde{\vh}_i = \frac{\vh_i - \bm\mu_s}{\|\vh_i - \bm\mu_s\|_F}$ where $\bm\mu_s$ is the average internal states, resulting in the feature matrix $\widetilde{\mathbf{S}}$.
We apply the same procedures on the target domain and obtain feature matrix $\widetilde{\mathbf{T}} \in \mathbb{R}^{M \times d}$. We rename $\widetilde{\mathbf{S}}$ to be ${\mathbf{S}}$ and $\widetilde{\mathbf{T}}$ to be ${\mathbf{T}}$ for simplicity. We first create independent $d^\prime$-dimension subspace for ${\mathbf{S}}$ and ${\mathbf{T}}$ respectively, to \textbf{preserve hallucination information}. Specifically, we first calculate the unbiased estimation of the covariance of $\mathbf{S}$ as 
$\bm\Sigma_{\mathbf{S}}=\frac{\mathbf{S}^T \cdot \mathbf{S}}{N-1}$, and conduct eigenvalue decomposition: 
\begin{equation}
\bm\Sigma_\mathbf{S} = \mathbf{Q}_{\mathbf{S}} \, \textit{diag}(\bm\Lambda_{\mathbf{S}}) \, \mathbf{Q}^T_{\mathbf{S}}
\end{equation}
for its eigenvalues $\bm\Lambda_{\mathbf{S}}$ and eigenvectors $\{\vk_j = \mathbf{Q}_{\mathbf{S}, :, j}\}_i^d$. Then, the independent subspace of $\mathbf{S}$ is created by spanning the eigenvectors corresponding to the top-$d^\prime$ eigenvalues, i.e., 
$\mathbf{K}_{\mathbf{S}} = \{\vk_1, \vk_2, \cdots, \vk_{d^\prime}\} \in \mathbb{R}^{d \times d^\prime}$. We apply the same procedures over $\mathbf{T}$ and obtain its independent subspace spanned by $\mathbf{K}_{\mathbf{T}}$. Since eigenvectors capture the directions with the greatest variance, the hallucination information encoded in $\mathbf{S}$ and $\mathbf{T}$ are preserved by $\mathbf{K}_{\mathbf{S}}$ and $\mathbf{K}_{\mathbf{T}}$, respectively.

\noindent\textbf{Distribution Alignment.} We further capture correlations $\mathbf{M}=\mathbf{K}_{\mathbf{S}}^T \cdot \mathbf{K}_{\mathbf{T}}$ to obtain the alignment matrix $\mathbf{M}$ for transiting from subspace $\mathbf{K}_{\mathbf{S}}$ to $\mathbf{K}_{\mathbf{T}}$ and apply it over $\mathbf{K}_{\mathbf{S}}$ to obtain aligned subspace $\mathbf{K}_{\mathbf{S}}^{\text{align}}= \mathbf{K}_\mathbf{S} \cdot \mathbf{M}$. Eventually, we project internal states via
\begin{equation}
\begin{aligned}
\small
    \bar{\vh}^T = \vh^T \cdot \mathbf{K}_{\mathbf{S}}^{\text{align}} & , \vh \sim \mathcal{S}, \\
    \bar{\vh}^T = \vh^T \cdot \mathbf{K}_{\mathbf{T}} & , \vh \sim \mathcal{D}, \\
\end{aligned}
\end{equation}
to make projected internal states well aligned. Denoting by $\mathcal{S}^\prime$ and $\mathcal{D}^\prime$ the aligned data domains, the hallucination detector is trained on $\mathcal{D}^\prime$ and evaluated on $\mathcal{S}^\prime$.

To be practical in real-world scenarios, we consider both \textit{\textbf{data}} and \textit{\textbf{model transferability}} at the same time, i.e., \textbf{co-transferring}: \textit{\textbf{(i)}} training hallucination detector on the \alignmethodname -aligned internal states collected from LVLM$_A$ over the training set of crafted CC-Sbu-Align hidden state dataset; \textit{\textbf{(ii)}} obtaining the thresholds $T_{\alpha \,fpr}$ which results FPR=$\alpha$ on the validation set of CC-Sbu-Align hidden state dataset; \textbf{\textit{(iii)}} testing the detector with thresholds $T_{\alpha \,fpr}$ on LVLM$_B (A\neq B)$ over the COCO 2014val dataset (we follow the same pipeline as in~\cref{sec:classifier} to collect internal states). For instance, the ``MiniGPT-4 $\rightarrow$ Llava-1.5'' plot (top left) in~\cref{fig:comnhallu} (b) indicates training a hallucination detector on the internal states collected from MiniGPT-4 over CC-Sbu-Align, and testing on Llava-1.5 over the COCO val2014.
We show that \alignmethodname effectively mitigates domain shifting and maintains high-specificity detection on various testing domains.
\section{\methodname: Truthful-Guided Decoding}\label{sec:intervention}

In this section, we demonstrate how to reduce OH during LVLM decoding under the guidance of truthful direction.

\subsection{Preliminary}
\begin{figure}[h]
    \centering
    \includegraphics[width=0.95\linewidth]{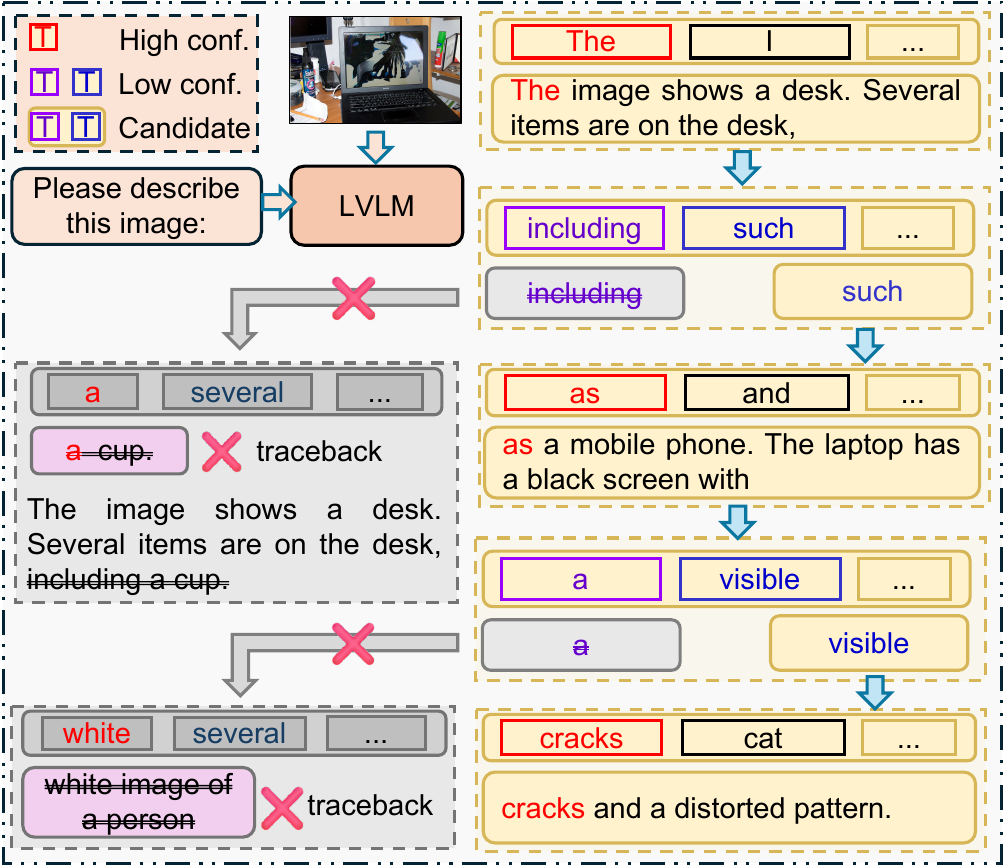}
    \caption{The schematic diagram of \methodname. When a hallucinated object token (e.g., ``cup'' for the first time) is detected, we trace it back by locating the token with the lowest confidence preceding this sentence (e.g., ``including'') and selecting the second candidate (e.g., ``such''). This process is repeated $\mathcal{N}_{B}$ times.}
    \label{fig:method}
    \vspace{-5mm}
\end{figure}

Given hallucination detector $\mathcal{G}$ trained in~\cref{sec:detection}, 
to mitigate hallucinations while preserving high-quality generation, it is essential to identify tokens that \textbf{\textit{(i)}} are close to the truthful domain and \textbf{\textit{(ii)}} maintain utility, e.g., minimal semantic distance to the input image for image caption task. Formally, this can be defined as:
\begin{equation*}
\small
    \arg\min_{\vz} \sum_{\{i | z_i \in \mathbb O\}} \mathds{1}[\mathcal G^*(\vh_{i-1})]+ d(\vx,z_{i\le n}, \vs),
\end{equation*}

\noindent where $\mathbb O$ represents the index set of objects token, and $d$ denotes the semantic distance metric between $\vx$ and $\vz$ following prompt $\vs$. To identify tokens with minimal distance to the image, we propose to pre-intervene model outputs with lower confidence scores when the optimal classifier $\mathcal G^*$ identifies potential hallucination behaviors and guides us on the need for constructing new tokens. In~\cref{fig:method}, a detailed diagram is provided to describe this procedure.

\subsection{Pre-Intervention: Motivation and Methods}\label{sec:motivation}

Specifically, we observed that
\ding{202} \textbf{the root cause of hallucinations may lie before the hallucinated token itself.} While hallucinations are typically detected in association with specific objects, the underlying triggers of these hallucinations may not be limited to the locations of the hallucinated objects~\cite{flemings2024characterizing,desrochers2024reducing}. For instance, consider an image that depicts only a ``\textit{dog}''. If the model generates the sentence: \{\emph{The image shows a dog running to the house.}\}, the phrase \{\emph{the house}\} constitutes a hallucinated object. However, the root cause of this hallucination might be attributed to the word \{\emph{to}\}. We further illustrate this in the bottom of Fig. \ref{fig:overall}. The inclusion of \{\emph{to}\} necessitates a subsequent noun for the sentence to feel complete, which may lead the model to hallucinate an object. 
Based on this insight, denote $e_{\vz}$ to be the index of first hallucination token in sequence $\vz$. We propose that upon detecting a hallucinated object, we first investigate whether any preceding token before $e_z$ within the sentence could have prompted the model to generate this hallucination. 

\begin{table*}[!htp]\centering
\adjustbox{width=0.95\textwidth}{
\begin{tabular}{lccccccccc}\toprule
 &\multicolumn{3}{c}{\textbf{MiniGPT-4}} &\multicolumn{3}{c}{\textbf{Llava-v1.5}} &\multicolumn{3}{c}{\textbf{mPlug-Owl2}} \\
\cmidrule(lr){2-4}
\cmidrule(lr){5-7}
\cmidrule(lr){8-10}
\textbf{Methods} &\textbf{CHAIR$_S\downarrow$} &\textbf{CHAIR$_I\downarrow$} &\textbf{BLEU$\uparrow$} &\textbf{CHAIR$_S\downarrow$} &\textbf{CHAIR$_I\downarrow$ }&\textbf{BLEU$\uparrow$} &\textbf{CHAIR$_S\downarrow$} &\textbf{CHAIR$_I\downarrow$ }&\textbf{BLEU$\uparrow$}  \\
\midrule
Greedy & 29.53$\pm$1.51 & 11.73$\pm$0.46 & 15.58$\pm$0.35 & 19.60$\pm$1.64 & 6.07$\pm$0.58 & 16.97$\pm$0.16 & 23.60$\pm$0.87 & 8.57$\pm$0.38 & 16.45$\pm$0.19 \\ 
 Beam Search & 25.80$\pm$0.00 & 10.15$\pm$0.21 & 16.06$\pm$0.37 & 19.40$\pm$1.70 & 6.55$\pm$0.92 & 17.24$\pm$0.23 & 19.90$\pm$0.42 & 7.30$\pm$0.42 & 16.69$\pm$0.12 \\ 
 DoLA & 26.00$\pm$1.41 & 10.25$\pm$0.35 & 16.05$\pm$0.39 & 18.60$\pm$3.39 & 6.35$\pm$1.20 & 17.18$\pm$0.28 & 20.20$\pm$0.28 & 7.45$\pm$0.21 & 16.81$\pm$0.16 \\ 
LURE & 27.88$\pm$2.25 & 10.20$\pm$0.85 &15.03$\pm$0.11 &19.48$\pm$2.35& 6.5$\pm$0.38& 15.97$\pm$0.01 &21.27$\pm$0.06 &7.67$\pm$0.16 & 15.65$\pm$0.05 \\
VCD & 28.93$\pm$2.47 & 12.10$\pm$0.79 & 15.18$\pm$0.63 & 23.00$\pm$2.95 & 7.47$\pm$0.50 & 15.78$\pm$0.13 & 24.80$\pm$1.51 & 9.07$\pm$0.91 & 15.43$\pm$0.18 \\ 
Woodpecker & 28.87$\pm$2.20 & 10.20$\pm$0.85 & 15.30$\pm$0.01 & 23.85$\pm$4.62 & 7.50$\pm$0.01&  17.05$\pm$0.00 & 26.33$\pm$1.98 & 8.43$\pm$0.80 & 16.43$\pm$0.00 \\
OPERA & 27.80$\pm$1.70 & 10.80$\pm$0.57 & 16.03$\pm$0.35 & 18.60$\pm$3.96 & 6.15$\pm$1.20 & 17.27$\pm$0.18 & 19.50$\pm$2.40 & 7.55$\pm$1.20 & 16.59$\pm$0.16 \\ 
HACL & 24.47$\pm$1.01 & 9.57$\pm$0.31 & 15.84$\pm$0.36 & 18.27$\pm$1.14 & 5.90$\pm$0.52 & 17.09$\pm$0.19 & 21.60$\pm$0.69 & 7.73$\pm$0.15 & 16.62$\pm$0.21 \\ 
Nullu & 21.40$\pm$1.00 & 8.99$\pm$0.36 & 14.81$\pm$0.06 & 15.20$\pm$0.60 & 5.30$\pm$0.03 & 15.69$\pm$0.04 & 15.60$\pm$1.20 & 5.77$\pm$0.01 & 15.45$\pm$0.01 \\
 \midrule
\rowcolor{LimeGreen!25}
\textbf{\methodname} & \textbf{16.87$\pm$0.87}&\textbf{7.53$\pm$0.33}&\textbf{17.21$\pm$0.75}&\textbf{10.33$\pm$3.31}&\textbf{3.87$\pm$1.16}&\textbf{19.79$\pm$0.15}&\textbf{11.13$\pm$1.50}&\textbf{5.27$\pm$0.42}&\textbf{18.82$\pm$0.22}\\
 
\bottomrule
\end{tabular}
}
\caption{The evaluation results on the COCO CHAIR benchmark. Lower CHAIR$_S$ and CHAIR$_I$ indicate fewer hallucinated objects. It is shown that \methodname significantly outperforms all the baselines in OH mitigation while resulting in higher-quality captions. }\label{tab:chair_results}
\vspace{-2mm}
\end{table*}

\begin{table*}
\parbox{0.6\linewidth}{
\centering
\adjustbox{width=\linewidth}{
    \begin{tabular}{lcccccc}\toprule
&\multicolumn{2}{c}{\textbf{MiniGPT-4}} &\multicolumn{2}{c}{\textbf{Llava-1.5}} &\multicolumn{2}{c}{\textbf{mPlug-Owl2}} \\
\cmidrule(lr){2-3}
\cmidrule(lr){4-5}
\cmidrule(lr){6-7}
\textbf{Methods}  &\textbf{Precision$\uparrow$} & $F_\beta\uparrow$ &\textbf{Precision$\uparrow$} & $F_\beta\uparrow$ &\textbf{Precision$\uparrow$} & $F_\beta\uparrow$ \\
\toprule
Greedy&90.13$\pm$1.19 & 88.86$\pm$1.15 & 92.80$\pm$1.08 & 91.73$\pm$1.13 & 91.39$\pm$0.72 & 90.30$\pm$0.68\\

Beam Search&91.57$\pm$0.11 & 90.22$\pm$0.17 & 93.10$\pm$0.40 & 91.99$\pm$0.31 & 92.12$\pm$0.56 & 90.86$\pm$0.57\\

VCD&89.85$\pm$0.97 & 88.49$\pm$0.83 & 92.33$\pm$1.08 & 91.29$\pm$1.04 & 90.74$\pm$0.40 & 89.57$\pm$0.37\\

OPERA&91.31$\pm$0.16 & 89.97$\pm$0.11 & 92.66$\pm$1.06 & 91.56$\pm$1.06 & 91.20$\pm$0.42 & 89.91$\pm$0.49\\

DoLA&91.92$\pm$0.31 & \textbf{90.56$\pm$0.26} & 93.13$\pm$0.40 & 92.02$\pm$0.34 & 91.92$\pm$0.37 & 90.67$\pm$0.39\\

HACL&91.21$\pm$1.27 & 89.75$\pm$1.22 & 92.62$\pm$0.87 & 91.53$\pm$0.88 & 91.26$\pm$0.56 & 90.08$\pm$0.56\\
\midrule
\rowcolor{LimeGreen!25}
\textbf{\methodname} & \textbf{92.28$\pm$1.28} & 90.03$\pm$1.21 & \textbf{94.28$\pm$0.60} & \textbf{92.47$\pm$0.60} & \textbf{93.66$\pm$0.73 }& \textbf{91.66$\pm$0.87} \\

\bottomrule
\end{tabular}
}
    \caption{Evaluation results on the offline POPE benchmark. Results are averaged over three splits (\textit{Random}, \textit{Popular}, and \textit{Adversarial}).}\label{tab:opope_result_average}
}
\hfill
\parbox{.37\linewidth}{
\centering
\adjustbox{width=\linewidth}{
    \begin{tabular}{ccccc}
      \toprule
      \textbf{LVLMs} & \textbf{Methods} & \textbf{CHAIR$_S\downarrow$} & \textbf{CHAIR$_I\downarrow$} & \textbf{BLEU$\uparrow$} \\
      \toprule
      \multirow{4}{*}{\makecell[c]{QWen2-VL-\\7B-Instruct\\(QWen2)}} & Greedy & 12.0 & 4.9 & 15.73 \\
      & Beam & 11.6 & 4.3 & 15.60 \\
      & HALC & 9.2 & 4.0 & 16.10 \\
      \cmidrule{2-5}
      & \cellcolor{LimeGreen!25}\textbf{\methodname} & \cellcolor{LimeGreen!25}\textbf{6.2} & \cellcolor{LimeGreen!25}\textbf{3.4} & \cellcolor{LimeGreen!25}\textbf{17.83} \\
      \midrule
      \multirow{4}{*}{\makecell[c]{InternVL-\\2.5-8B\\(InternLM2)}} & Greedy & 13.2 & 4.8 & 17.43 \\
      & Beam & 11.8 & 4.6 & 17.70 \\
      & HALC & 10.1 & 4.2 & 17.90 \\
      \cmidrule{2-5}
      & \cellcolor{LimeGreen!25}\textbf{\methodname} & \cellcolor{LimeGreen!25}\textbf{3.2} & \cellcolor{LimeGreen!25}\textbf{3.0} & \cellcolor{LimeGreen!25}\textbf{18.26} \\
      \bottomrule
    \end{tabular}
    }
    \caption{Evaluating the transferability to advanced LVLMs using backbones \textit{other than Llama} with mis-matched dimensionality.}\label{tab:non_llama_transfer}
}
  \vspace{-5mm}
\end{table*}

To locate the preceding token triggering hallucination, we analyze LVLM output confidences and observe that \ding{203} \textbf{tokens with lower confidence frequently precede hallucinated objects} (please refer to~\cref{appendix:preceding_token_confidence} for more experimental evidence). This aligns with the idea that some hallucinations arise from ambiguous information provided to the model or its inability to respond appropriately~\cite{hou2023decomposing,huang2024survey}, leading it to select an incorrect or irrelevant word. When the model is uncertain about how to proceed, its confidence in generating a response decreases significantly.

Building on this observation, we propose the following approach: Let $\mathbf{o}^k_i$ represent the confidence score of location $i$ from the LVLM $\mathcal{M}$’s output after applying softmax following the $k$-th backtrace, where $\vo_{i}^{k} = \mathcal{M}^{o}(\vx, \vs, \vz^k_{<i};\bm\theta)$, and $\text{TopK}(\mathbf{o}^k_i, 1)$ is the largest confidence candidate in $\mathbf{o}^k_i$. When identifying trigger words, we begin with the hallucinated token $e_{\vz}$ and move backward through the sentence to locate the token $i$ with the lowest top confidence $\text{TopK}(\mathbf{o}^k_i, 1)$. 

From these candidates, we exclude previously selected tokens and choose a new one. 
Denote $r_{z_i}$ to be the rank of the selected token in $\mathbf{o}^k_i$. To make a selection, we consider all possible candidates suggested by the model. Specifically, we rank these candidates in descending order of likelihood and select the highest-ranked token $\text{TopK}(\mathbf{o}^k_i, r_{z_i} + 1)$ at each iteration. This strategy is reasonable in scenarios without supplementary tools, such as additional LLVMs. 
However, this method does not guarantee that the second choice will be the correct trigger word. Consequently, after selecting a new token, we repeat this process iteratively. When tracing back to identify the trigger word, it is crucial to limit the search to tokens located within a relatively short distance from the hallucinated token, as the causal relationship diminishes with increasing distance. Naturally, this search is constrained to the sentence in which the hallucination occurs. 

Finally, we consider the hallucinated word itself. The previously outlined methods are not entirely reliable in pinpointing the precise trigger word, meaning our algorithm may continue detecting hallucinations even after several iterations. In such cases, persisting with the above algorithm is suboptimal: not only may the identified words fail to represent the actual triggers, but the number of candidate tokens suggested by the model is limited. 
To address this issue, when further iterations are unlikely to yield results, we set the max number of backtrace to be $\mathcal N_B$, and opt to select the second candidate $\text{TopK}(\mathbf{o}^k_i, 2)$ provided by the model for the corresponding hallucinated word after achieving $\mathcal N_B$. \cref{appendix:procedure} provides an outline of the proposed method (for simplicity, we define FindFirstHallucination($\vz$) as the process of using $\mathcal{G}$ to identify the first hallucinated token of $\vz$ and output the index, and define the classifier $\mathcal G(\vh_i^k) < \tau,$ if $i + 1 \notin \{i|z_i\in \mathcal O\}$).
\section{Experiments}

\noindent\textbf{Benchmarks and Baselines.} We follow previous work and evaluate our methods on popular OH benchmarks, including MSCOCO CHAIR evaluation~\cite{rohrbach2018object}, POPE~\cite{li2023evaluating}, Offline POPE~\cite{chenhalc}, and qualitative examination on LLaVA-bench~\cite{liu2024improved}. Please refer to~\cref{appendix:benchmarks} for the introduction of each benchmark. We consider 8 competitive baselines, including naive Greedy generation, Beam search (with beams set to 3), VCD~\cite{leng2024mitigating}, OPERA~\cite{huang2024opera}, DoLA~\cite{chuangdola}, HALC~\cite{chenhalc}, Woodpecker~\cite{yin2023woodpecker}, LURE~\cite{zhouanalyzing}, and Nullu~\cite{yang2024nullu}. We follow the original hyperparameters of each baseline according to their papers or codebases.

\noindent\textbf{LVLMs and Co-Transferring Settings.} We consider three advanced LVLMs, including MiniGPT-4~\cite{zhuminigpt}, Llava-1.5~\cite{liu2024visual}, and mPlug-Owl2~\cite{ye2024mplug}. For LVLMs with non-Llama backbones, we consider the powerful Qwen2-VL-7B-Instruct~\cite{wang2024qwen2} and InternVL-2.5-8B~\cite{chen2024expanding}. To be practical in the real world, we apply the co-transferring settings as we mentioned in~\cref{sec:comnhallu_transfer}: (i) training the hallucination detector with \alignmethodname -aligned MiniGPT-4 internal states over the crafted CC-Sbu-Align dataset and obtaining thresholds $T$ that result in FPR=$\alpha$ on the validation set; (ii) testing the detector with the threshold $T$ on other LVLMs.

\noindent\textbf{Default Hyperparameters.} For all the experiments, we set threshold $\tau$ to be 0.4, the subspace dimension in \alignmethodname, i.e., $d^\prime$, to be 64, the layer index for hidden states collection $l$ to be middle layer 16, and the maximum allowed traceback times to be 5. Following~\cite{chenhalc}, we randomly select 500 images for each experiment and repeat three times, reporting both average performance and standard derivations. In~\cref{sec:ablationstudy}, we provide detailed ablation studies for hyperparameters. We utilize the prompt ``\textit{Please describe this image in detail.}'' for all caption generation.

\begin{figure}
    \centering
    \includegraphics[width=0.93\linewidth]{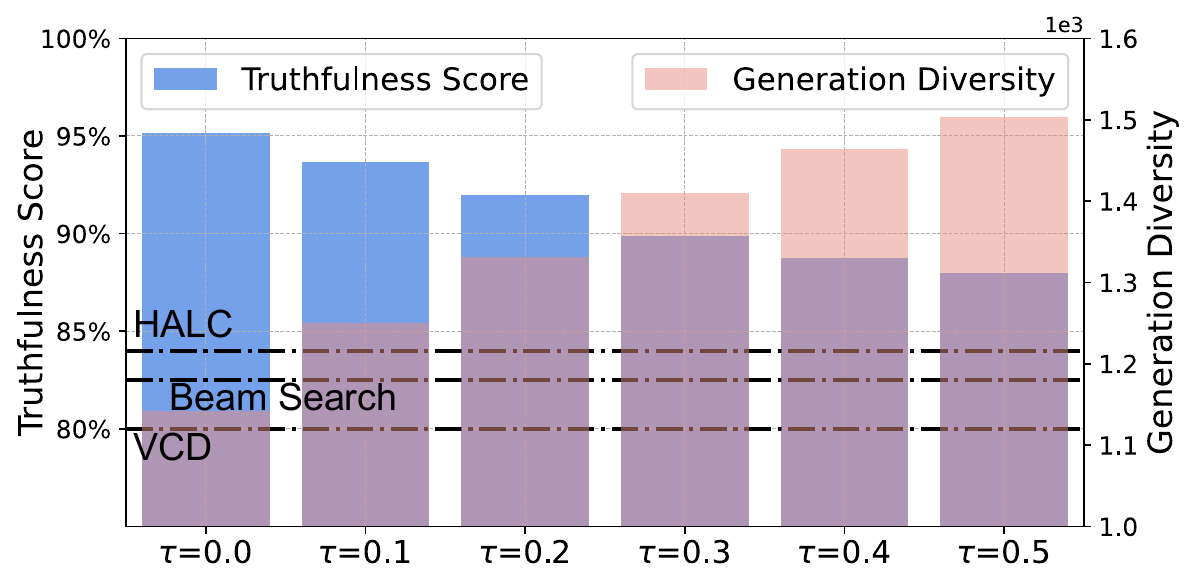}
    \caption{Trade-off between truthfulness and diversity. We show that \methodname offers flexible adjusting of threshold $\tau$: smaller $\tau$ for truthfulness in safety-critical scenarios while larger $\tau$ for diverse generations.}
    \label{fig:truthful_divers_tradeoff}
    \vspace{-5mm}
\end{figure}

\subsection{CHAIR Evaluation.}
In~\cref{tab:chair_results}, we report CHAIR$_S$ for the portion of hallucinated captions, CHAIR$_I$ for the portion of hallucinated objects, and the quality of generated captions measured by BLEU~\cite{papineni2002bleu}. It is shown that our method significantly outperforms all the baselines in both OH mitigation and caption quality. Specifically, \methodname outperforms the current state-of-the-art method HALC by 12\% to 14\% CHAIR$_S$ and over 2\% CHAIR$_I$ over all three LVLMs. Moreover, \methodname substantially improves the quality of captions where it outperforms baselines by nearly 2\% BLEU across all the settings, suggesting that truthful guidance not only mitigates the hallucination behaviors of LVLM but also enables high-quality caption generation. 

\noindent \textbf{Non-Llama Backbone and Mis-matched Dimensionality} In~\cref{tab:non_llama_transfer}, we provide the transferability evaluation over the non-Llama LVLMs. It is shown that \methodname demonstrates significant transferability when transferring MiniGPT-4 (Vicuna as backbone LLM) hidden states to various LLM backbones, e.g., QWen2-VL (QWen2 as backbone LLM) and InternVL-2.5 (InternLM2 as backbone LLM). Moreover, in terms of dimension mismatch, \alignmethodname incorporates a subspace projection mechanism to standardize hidden state dimensions, which could be applied in addition to handling dimension mismatch. The experiment of MiniGPT-4 (4,096 dimensions) $\rightarrow$ QWen2VL-7B-Instruct (3,588 dimensions) supports the flexibility and transferability of our design.

\subsection{POPE Evaluation.}
As highlighted in previous work~\cite{chenhalc}, the original POPE benchmark requires robust chat capability to LVLMs for question answering. We follow~\cite{chenhalc} to conduct offline POPE (OPOPE), where we derive questions and answers from LVLM descriptions. We use Precision and $F_\beta$ metric with $\beta=0.1$ for overall performance comparison. The averaging results are summarized in~\cref{tab:opope_result_average} (the full results, as well as the original POPE evaluation results, are provided in~\cref{appendix:pope_results}). It is shown that \methodname achieves the best Precision among most settings and splits, indicating that the high-specificity design works well.

\begin{figure}[t]
    \centering
    \includegraphics[width=0.95\linewidth]{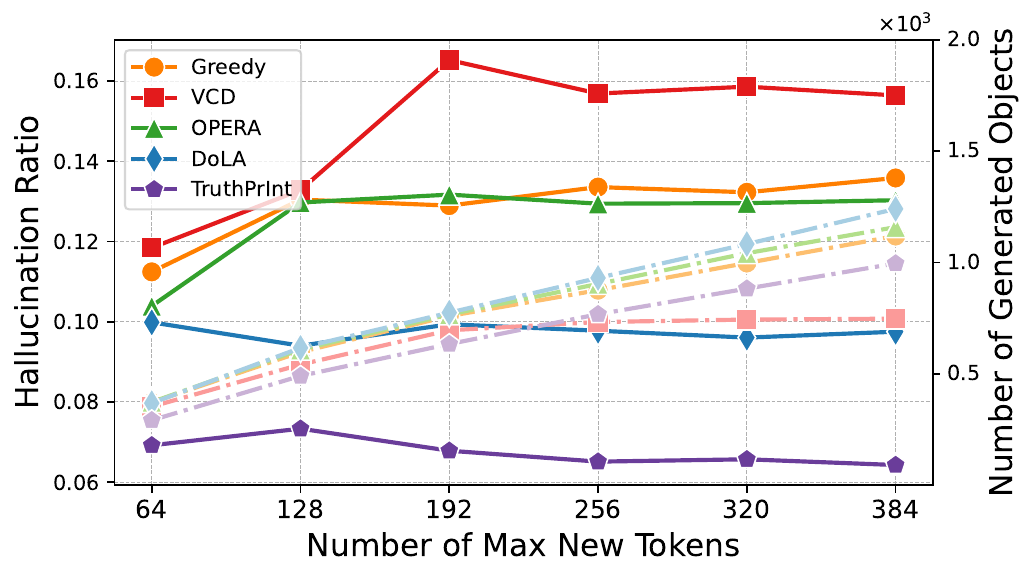}
    \vspace{-3mm}
    \caption{Hallucination ratio and number of generated objects under various ``maximum new token'' limitations. }
    \label{fig:longer_captions}
    \vspace{-3mm}
\end{figure}

\begin{table}
    \centering
    \adjustbox{width=0.95\linewidth}{
    \begin{tabular}{cccccc}
    \toprule
         \textbf{\makecell{$\mathcal N_B$}} & \textbf{CHAIR$_S \downarrow$} & \textbf{CHAIR$_I\downarrow$} & \textbf{BLEU$\uparrow$} & \textbf{Precision$\uparrow$} & \textbf{$F_\beta \uparrow$} \\
        \midrule
        1 & 16.20 & 7.70 & 17.60 & 92.73 & 90.86\\
2 & 16.00 & 7.40 & 17.58 & 93.50 & 91.58\\
3 & 15.20 & 6.90 & 17.55 & 93.91 & 91.92\\
4 & 15.60 & 7.00 & 17.45 & 93.57 & 91.58\\
5 & 15.40 & 7.10 & 17.43 & 93.46 & 91.48\\
        \bottomrule
    \end{tabular}
    }
    \caption{Ablation study on the maximum number of traceback. Enabling more $\mathcal N_B$ allows more ``trial and error'' to remove OH.}
    \label{tab:ablationstudy_N_B}
    \vspace{-4mm}
\end{table}

\begin{figure*}
    \centering
    \includegraphics[width=0.85\linewidth]{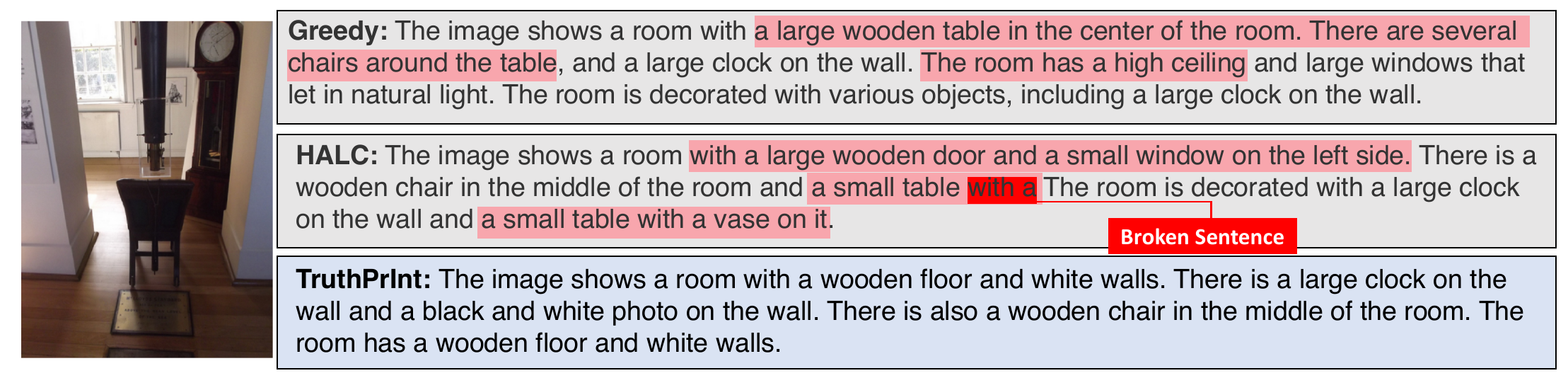}
    \caption{Qualitative analysis of generated captions. Both Greedy search and HALC encode lots of hallucinated objects, such as ``\textbf{table}'', ``\textbf{ceiling}'', ``\textbf{vase}'', etc. Moreover, HALC experiences \textbf{broken sentences} due to token replacement. \methodname provides detailed and accurate descriptions, even including the small object –``\textbf{black and white photo on the wall}''.}
    \label{fig:quantitative_analysis}
    \vspace{-6mm}
\end{figure*}

\subsection{Ablation Study}\label{sec:ablationstudy}
We perform ablation studies on MiniGPT-4 using the COCO val2014 dataset without further specification. During the study, all hyperparameters remain the same as default values except for the parameter being ablated.

\noindent\textbf{Trade-off: Truthful or Diverse?} Threshold $\tau$ decides the criterion of hallucination identification: smaller $\tau$ means a lower standard for hallucination (or higher standard for truthful) identification, enabling more tokens to be regarded as hallucinated. Inevitably, this will reject substantial decoding trajectories and conflict with generation quality, especially diversity. To quantify this, we investigate the relationship between \textit{truthfulness score}: $(100 - \frac{\text{CHAIR}_S + \text{CHAIR}_I}{2})$ and generation diversity measured by the number of objects generated. As shown in~\cref{fig:truthful_divers_tradeoff}, we suggest adjusting $\tau$ according to application scenarios, e.g., smaller $\tau$ in safety-critical scenarios to embrace more truthfulness.

\noindent\textbf{OH in Longer Captions.} Recent work reveals that it is essential to evaluate OH mitigation in longer captions since \textit{(i)} OH happens more frequently in longer captions with more objects mentioned~\cite{zhou2023analyzing,huang2024opera}; \textit{(ii)} it will not hurt natural performance, e.g., providing high-quality and diverse generations. In~\cref{fig:longer_captions}, we report hallucination ratios, i.e., how many generated objects are hallucinated, and the total number of generated objects. It is shown that \methodname exhibits significantly low hallucination ratios when generating longer captions while maintaining close object numbers.

\begin{table}
    \centering
    \adjustbox{width=0.97\linewidth}{
    \begin{tabular}{cccccc}
    \toprule
        \textbf{Layer $l$ }  & \textbf{CHAIR$_S\downarrow$} & \textbf{CHAIR$_I\downarrow$} & \textbf{BLEU$\uparrow$} & \textbf{Precision$\uparrow$} &\textbf{$F_\beta \uparrow$} \\
        \midrule
        Greedy & 29.53 & 11.73 & 15.58 & 90.12 & 88.85 \\
        \midrule
12 & 24.20 & 10.70 & 17.80 & 91.76 & 90.31\\
14 & 10.40 & 4.90 & 17.57 & 94.90 & 92.70\\
16 & 15.40 & 7.10 & 17.43 & 93.46 & 91.48\\
18 & 10.00 & 5.30 & 17.35 & 94.69 & 92.21\\
20 & 11.80 & 6.50 & 17.20 & 93.60 & 91.13\\
        \bottomrule
    \end{tabular}
    }
    \caption{Ablation study on the layers of LVLM for internal states collection and hallucination detection.}
    \label{tab:ablation_layer_idx}
    \vspace{-6mm}
\end{table}

\begin{figure}
    \centering
    \includegraphics[width=0.97\linewidth]{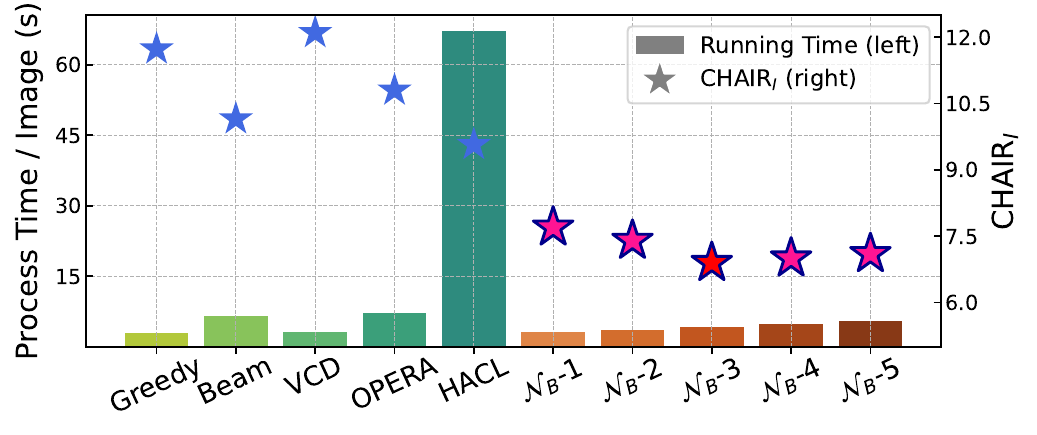}
    \caption{Efficiency comparison. \methodname requires similar computational costs as Greedy search while achieving better performance. $\mathcal{N}_B$ substantially boosts OH mitigation involving limited computational overhead.}
    \label{fig:efficiency_comparison}
    \vspace{-6mm}
\end{figure}

\noindent\textbf{Efficiency and Number of Tracebacks.} 
Unlike existing post-processing methods where heavy auxiliary models, e.g., LLMs~\cite{yin2023woodpecker} and CLIP~\cite{ouali2024clip}, are incorporated. \methodname leverages simple MLP models and limited backtracking mechanisms for truthful guidance, which exhibit close efficiency to naive Greedy search. In~\cref{fig:efficiency_comparison}, we present the per-image process time consumed by baselines and \methodname. For \methodname, we included the \textbf{detector training overhead} and provided the efficiency under various maximum numbers of tracebacks $\mathcal N_B$. Results are obtained by averaging MiniGPT-4 over 500 images on a single A40 GPU. It is shown that \methodname requires close computational costing as Greedy yet achieves significant improvements. Also, enlarging $\mathcal{N_B}$ substantially reduces OH from 7.7 CHAIR$_I$ to around 7.0 (~\cref{tab:ablationstudy_N_B}), meaning the designed backtracking is efficient and effective.

\noindent\textbf{Internal States Layer $l$.} We investigate which layers $l$ of hidden states in LVLMs more effectively encode truthfulness information. In~\cref{tab:ablation_layer_idx}, we show that middle layers typically encode more truthfulness of generations~\cite{chen2024inside,du2024haloscope}.

\subsection{Qualitative Analysis}
We manually examine the quality of generated captions on COCO val2014 and LLaVA-Bench~\cite{liu2024improved} (\cref{appendix:llavabench}). In~\cref{fig:quantitative_analysis}, we present one of the captions generated by Greedy search, HALC, and \methodname regarding the same image. It is shown that \methodname provides more accurate and detailed descriptions than baselines.

\vspace{-1mm}
\section{Conclusion}
In this paper, we investigate OH in LVLMs, which is one of the most serious trustworthy issues. Our research starts with the discovery that LVLM internal states, e.g., hidden states, are high-specificity and transferrable hallucination indicators. Based on that, we propose \methodname, which first learns truthful direction in latent space and then applies truthful-guided intervention for OH mitigation during testing time. Our work highlights that internal states encode per-token truthfulness information. Extensive results show that \methodname outperforms existing baselines with significant margins. 

\section*{Acknowledgment}
This work was performed under the auspices of the U.S. Department of Energy by Lawrence Livermore National Laboratory under Contract DE-AC52-07NA27344 and LDRD
Program Project No.~23-ERD-030 (LLNL-JRNL-2003786).

{
    \small
    \bibliographystyle{ieeenat_fullname}
    \bibliography{main}
}

\clearpage
\setcounter{page}{1}
\maketitlesupplementary
\appendix

\section{Hallucination Detection with Internal States}
\subsection{Internal States Collection}\label{appendix:collect_hs}
In~\cref{sec:classifier}, we utilize the hidden states of preceding tokens associated with object tokens to detect hallucinations. Specifically, the hallucination detector is designed to provide an early warning by predicting whether future object tokens are likely to be hallucinated. This approach ensures that the detector is not exclusively trained on object tokens but functions as a generalized detector applicable to any type of token. From an intervention perspective, this ``early warning'' mechanism reduces the inference time of the LLM during decoding. For example, when determining the next token $z_j$, the previous hidden states can be directly passed to the detector for hallucination identification, i.e., $\mathcal{G}(\vh_{j-1}) < \tau$. In contrast, a ``current-token'' prediction approach would require computing the current hidden states $\vh_j$, which involves an additional LLM inference step before detecting hallucinations, i.e., $\mathcal{G}(\vh_{j}) < \tau$.

\subsection{Training Protocol of Hallucination Detection}\label{appendix:classifier_training}
In our implementation, the hallucination detector $\mathcal{G}$ is a 3-layer MLP, with the architecture presented in~\cref{tab:detector_arch}. The model is trained for 30 epochs with a batch size of 512, a learning rate of 0.001, and the Adam optimizer, utilizing binary cross-entropy (BCE) as the training objective. The optimal checkpoint is determined based on its performance on the validation set.

\begin{table}[h]
    \centering
    \adjustbox{width=\linewidth}{
    \begin{tabular}{c|c|c|c}
    \toprule
    \textbf{Layer 1} & \textbf{Layer 2} & \textbf{Layer 3} & \textbf{Activation} \\
    \midrule
      \texttt{(4096, 128)} & \texttt{(128, 64)} & \texttt{(64, 1)} & \texttt{ReLu}  \\
    \bottomrule
    \end{tabular}
    }
    \caption{The architecture of $\mathcal{G}$.}
    \label{tab:detector_arch}
\end{table}

\section{\methodname: Preliminary Analysis}
\subsection{Low-Confidence Tokens Precede Hallucination }\label{appendix:preceding_token_confidence}
As we mentioned in~\cref{sec:motivation}, tokens with lower confidence frequently precede hallucinated objects. Here, we provide experimental evidence to support it.
Specifically, for each object token, we calculate \textit{Preceding Minimum Confidence} (\textbf{PMC}): the minimum LVLM confidence of the preceding tokens of the object token within the same sentence. In~\cref{tab:stat}, we present the average PMC collected from hallucinated object tokens and truthful object tokens, respectively, over 500 samples. 
It is shown that the PMC of hallucinated is significantly larger than the PMC of truthful object tokens, indicating that low-confidence tokens tend to derive hallucinated objects.

\begin{table}[h]
\centering
\adjustbox{width=\linewidth}{
\begin{tabular}{ccc}
\toprule
\textbf{Model}      & \textbf{PMC of Hallucinated} & \textbf{PMC of Truthful} \\ \midrule
MiniGPT-4   & 0.39          & 0.31      \\ 
Llava-1.5  & 0.29          & 0.22      \\ 
mPlug-Owl2 & 0.29          & 0.20      \\ \bottomrule
\end{tabular}
}

\caption{The average \textit{Preceding Minimum Confidence} (PMC) over hallucinated and truthful object tokens. The PMC of hallucinated objects is significantly larger than the PMC of truthful object tokens, indicating that tokens with lower confidence frequently preceded hallucinated objects.}
\label{tab:stat}
\end{table}

\begin{algorithm}[tbp]
\caption{\methodname decoding}
\label{algorithm1}
\begin{algorithmic}[1]
\State \textbf{Input:} Prompt $\vs$, model $\mathcal M$, the image $\vx$, max backtracing number $\mathcal N_B$, detector $\mathcal G$, target layer $L$, threshold $\tau$
\State $k=0,\, i=0$ 
\State $\vr=\mathbf 0$ \Comment{Rank of Selected Token}
\State $\vc=\mathbf 0 \in \mathbb N^{\mathcal N_B+1}$ \Comment{\# of Hallucination}
\Repeat
    \Repeat \Comment{Generate a Sentence}
    \State $\vo_i^k = \mathcal{M}^{o}(\vx, \vs, \vz^k_{<i};\bm\theta)$
    \State $\vh_{i-1}^k=\mathcal{M}^{L}(\vx, \vs, \vz^k_{<i};\bm\theta)$
    \State $z_i^k = \text{TopK}(\vo_i^k, \vr_i+1) $ \Comment{Next Rank Token}
    \State $\vc_k = \vc_k + \mathds 1[\mathcal G(\vh_i^k) > \tau]$
    \State $\vr_i=\vr_i+\mathds 1[\mathcal G(\vh_i^k) > \tau]$
    \State $i = i + 1$ \Comment{Generate Next Token}
\Until{$z^k_{i-1}$ in $[eos,.]$}
\If{$\vc_i = 0$} \Comment{No Hallucination}
\State\Return $\vz^k$
\Else \Comment{Next Backtracing Initialization}
\State $k = k + 1$ 
\State $i^k=\argmin(\{\text{TopK}(\vo_{j}^{k-1}, 1)| j \le i\})$
\State $\vz_{<i^k}^k=\vz_{<i^k}^{k-1}, \, i=i^k$ 
\State $\vr_{>i}=0$ \Comment{Set State and Backtracing From $i^k$}
\EndIf
\Until{$k>\mathcal N_B$} \Comment{Achieve the Max Backtracing Number}
\Comment{Find Sentence with Less Hallucination}
\State $k'=\argmin(\vc_{\le\mathcal N_B})$ 
\State $i=\text{FindFirstHallucination}(\vz^{k'})$
\State $\vz_{<i}^{k}=\vz_{<i}^{k'}$ \Comment{Backtracing from $i$}
    \Repeat
    \State $\vo_i^k = \mathcal{M}^{o}(\vx, \vs, \vz^k_{<i};\bm\theta)$
    \State $\vh_{i-1}^k=\mathcal{M}^{L}(\vx, \vs, \vz^k_{<i};\bm\theta)$
    \State $z_i^k = \text{TopK}(\vo_i^k, \mathds 1[\mathcal G(\vh_{i-1}^k) > \tau]+1) $
    \State $\vc_k = \vc_k + \mathds 1[\mathcal G(\vh_{i-1}^k) > \tau]$ 
    \State $i = i + 1$
\Until{$z^k_{i-1}$ in $[eos,.]$}
\State $k=\argmin(\vc)$
\State \Return $\vz^k$
\end{algorithmic}
\end{algorithm}

\subsection{Method Procedures}\label{appendix:procedure}
In~\cref{algorithm1}, we present our pre-intervention mechanism algorithmic descriptions.

\section{Experiment Protocols}
In this section, we introduce the OH benchmarks used in this paper and additional experimental results as well.
\subsection{Benchmarks}\label{appendix:benchmarks}
\noindent\textbf{MSCOCO CHAIR}~\cite{rohrbach2018object} is a widely used benchmark for evaluating OH. Given a set of images, it tasks LVLMs with generating detailed descriptions of the images. The next step involves comparing the objects present in the images with those mentioned by the LVLMs, using specific metrics 
$$
\text{CHAIR}_S = \frac{|\text{sentences with hallucinated objects}|}{|\text{all sentences}|} \\
$$
$$
\text{CHAIR}_I = \frac{|\text{hallucinated objects}|}{|\text{all objects mentioned}|}
$$
for OH evaluation. It is usually incorporated with the COCO image caption dataset.

\noindent\textbf{POPE}~\cite{li2023evaluating} conducts an empirical evaluation of OH across multiple LVLMs, revealing its severity and identifying critical factors influencing this issue. It introduces Polling-based Object Probing Evaluation (POPE), which reformulates hallucination assessment as a binary classification task to improve stability, fairness, and scalability over existing methods.

\noindent\textbf{LLaVA-Bench}~\cite{liu2024improved} is a diverse collection of 24 images featuring various contexts, such as in-door, and outdoor. Each image is paired with a meticulously crafted, detailed description and a thoughtfully chosen set of questions. It is usually used for quantitative analysis of LVLM behaviors.

\subsection{POPE Results}\label{appendix:pope_results}
In~\cref{tab:opope_results}, we present the individual results over each offline POPE split. We also provide the original POPE evaluation results, obtained from MiniGPT-4 for each split, in~\cref{tab:pope_results}.

\begin{table*}[!htp]\centering
\adjustbox{width=0.8\linewidth}{
\begin{tabular}{lcccccccc}\toprule
&\multicolumn{2}{c}{\textbf{Random}} &\multicolumn{2}{c}{\textbf{Popular}} &\multicolumn{2}{c}{\textbf{Adversarial}} &\multicolumn{2}{c}{\textit{average}} \\
\cmidrule(lr){2-3}
\cmidrule(lr){4-5}
\cmidrule(lr){6-7}
\cmidrule(lr){8-9}
\textbf{Method} &\textbf{Precision$\uparrow$} &$F_\beta\uparrow$ &\textbf{Precision$\uparrow$} &$F_\beta\uparrow$ &\textbf{Precision$\uparrow$} &$F_\beta\uparrow$ &\textbf{Precision$\uparrow$} &$F_\beta\uparrow$ \\\midrule
Greedy &67.65 &67.78 &55.60 &55.79 &58.97 &59.15 &60.74 &60.91 \\
VCD &60.76 &60.79 &52.63 &52.70 &54.33 &54.38 &55.91 &55.96 \\
Beam &64.30 &64.47 &54.68 &54.88 &56.44 &56.64 &58.47 &58.66 \\
\midrule
\rowcolor{LimeGreen!25}
\textbf{\methodname} &\textbf{68.23} &\textbf{68.35} &\textbf{55.76} &\textbf{55.93} &\textbf{59.09} &\textbf{59.26} &\textbf{61.03} &\textbf{61.18} \\
\bottomrule
\end{tabular}
}
\caption{Evaluation results on the original POPE benchamrk.}\label{tab:pope_results}
\end{table*}

\begin{table*}[!htp]\centering
\adjustbox{width=\linewidth}{
\scriptsize
\begin{tabular}{llcccccc}\toprule
& &\multicolumn{2}{c}{\textbf{MiniGPT4}} &\multicolumn{2}{c}{\textbf{Llava-1.5}} &\multicolumn{2}{c}{\textbf{mPlug-Owl2}} \\
\cmidrule(lr){3-4}
\cmidrule(lr){5-6}
\cmidrule(lr){7-8}

\textbf{POPE Split} & \textbf{Methods} &\textbf{Precision$\uparrow$} &$F_\beta\uparrow$ &\textbf{Precision$\uparrow$} &$F_\beta\uparrow$ &\textbf{Precision$\uparrow$} &$F_\beta\uparrow$ \\\midrule
\multirow{7}{*}{Random} &Greedy &97.13$\pm$0.22 &95.59$\pm$0.16 &98.21$\pm$0.16 &\textbf{96.95$\pm$0.06} &96.66$\pm$1.44 &\underline{95.39$\pm$1.46} \\
&Beam &97.51$\pm$0.92 &\underline{95.93$\pm$0.80} &97.70$\pm$0.14 &96.43$\pm$0.22 &96.47$\pm$1.76 &95.05$\pm$1.67 \\
&VCD &96.78$\pm$1.42 &95.14$\pm$1.35 &97.11$\pm$1.18 &95.91$\pm$1.06 &96.80$\pm$0.87 &\textbf{95.40$\pm$0.84} \\
&OPERA &98.12$\pm$0.51 &96.51$\pm$0.44 &97.70$\pm$0.46 &96.43$\pm$0.48 &96.10$\pm$1.30 &94.62$\pm$1.15 \\
&DOLA &97.51$\pm$0.52 &\textbf{95.94$\pm$0.46} &97.70$\pm$0.12 &96.43$\pm$0.17 &96.47$\pm$1.35 &95.04$\pm$1.27 \\
&HALC &97.04$\pm$0.39 &95.33$\pm$0.38 &97.98$\pm$1.01 &96.60$\pm$1.00 &96.73$\pm$1.24 &95.35$\pm$1.20 \\
\cmidrule{2-8}
\rowcolor{LimeGreen!25}
&\textbf{\methodname} &\textbf{98.17$\pm$0.46} &95.58$\pm$0.46 &\textbf{98.65$\pm$0.80} &\underline{96.63$\pm$0.86} &\textbf{97.48$\pm$0.64} &95.28$\pm$0.71 \\
\midrule
\multirow{7}{*}{Popular} &Greedy &87.50$\pm$2.16 &86.34$\pm$2.10 &91.63$\pm$1.32 &90.60$\pm$1.36 &89.69$\pm$1.36 &88.66$\pm$1.26 \\
&Beam &89.61$\pm$1.01 &\underline{88.34$\pm$1.04} &90.92$\pm$0.50 &89.88$\pm$0.41 &90.30$\pm$3.05 &89.12$\pm$2.97 \\
&VCD &87.12$\pm$0.87 &85.87$\pm$0.74 &91.11$\pm$1.69 &90.11$\pm$1.66 &89.18$\pm$0.46 &88.07$\pm$0.40 \\
&OPERA &88.85$\pm$0.84 &87.61$\pm$0.85 &90.52$\pm$2.19 &89.49$\pm$2.14 &89.42$\pm$1.22 &88.21$\pm$1.26 \\
&DOLA &90.13$\pm$0.19 &\textbf{88.85$\pm$0.22} &91.14$\pm$0.25 &90.09$\pm$0.19 &90.01$\pm$2.72 &88.83$\pm$2.65 \\
&HALC &89.16$\pm$1.51 &87.79$\pm$1.44 &90.90$\pm$1.10 &89.86$\pm$1.10 &89.50$\pm$1.10 &88.39$\pm$1.06 \\
\cmidrule{2-8}
\rowcolor{LimeGreen!25}
&\textbf{\methodname} &\textbf{90.23$\pm$1.66} &88.10$\pm$1.47 &\textbf{93.13$\pm$0.86} &\textbf{91.38$\pm$0.90} &\textbf{92.64$\pm$1.75} &\textbf{90.70$\pm$1.76} \\
\midrule
\multirow{7}{*}{Adversarial} &Greedy &85.75$\pm$1.53 &84.64$\pm$1.48 &88.56$\pm$2.07 &87.63$\pm$2.08 &87.82$\pm$1.79 &86.85$\pm$1.71 \\
&Beam &87.59$\pm$0.22 &86.40$\pm$0.28 &90.69$\pm$0.83 &\textbf{89.66$\pm$0.73} &89.58$\pm$0.40 &88.42$\pm$0.41 \\
&VCD &85.64$\pm$1.53 &84.45$\pm$1.38 &88.78$\pm$1.97 &87.85$\pm$1.92 &86.24$\pm$0.61 &85.23$\pm$0.58 \\
&OPERA &86.97$\pm$0.80 &85.80$\pm$0.73 &89.78$\pm$0.54 &88.77$\pm$0.55 &88.07$\pm$1.33 &86.91$\pm$1.36 \\
&DOLA &88.10$\pm$0.60 &\textbf{86.90$\pm$0.54} &90.56$\pm$1.07 &\underline{89.53$\pm$0.99} &89.29$\pm$0.26 &88.13$\pm$0.22 \\
&HALC &87.44$\pm$2.65 &86.13$\pm$2.54 &88.99$\pm$1.17 &88.02$\pm$1.17 &87.55$\pm$1.59 &86.50$\pm$1.55 \\
\cmidrule{2-8}
\rowcolor{LimeGreen!25}
&\textbf{\methodname} &\textbf{88.44$\pm$2.09} &\underline{86.42$\pm$2.05 } &\textbf{91.06$\pm$1.19} &89.40$\pm$1.05 &\textbf{90.86$\pm$1.82} &\textbf{89.01$\pm$1.82} \\
\bottomrule
\end{tabular}
}
\caption{Evaluation results of each offline POPE split.}\label{tab:opope_results}
\end{table*}

\subsection{LLaVA-Benchmark Quantitative Analysis}\label{appendix:llavabench}
\begin{figure*}[ht]
        \begin{subfigure}[b]{1\textwidth}
                \includegraphics[width=\linewidth]{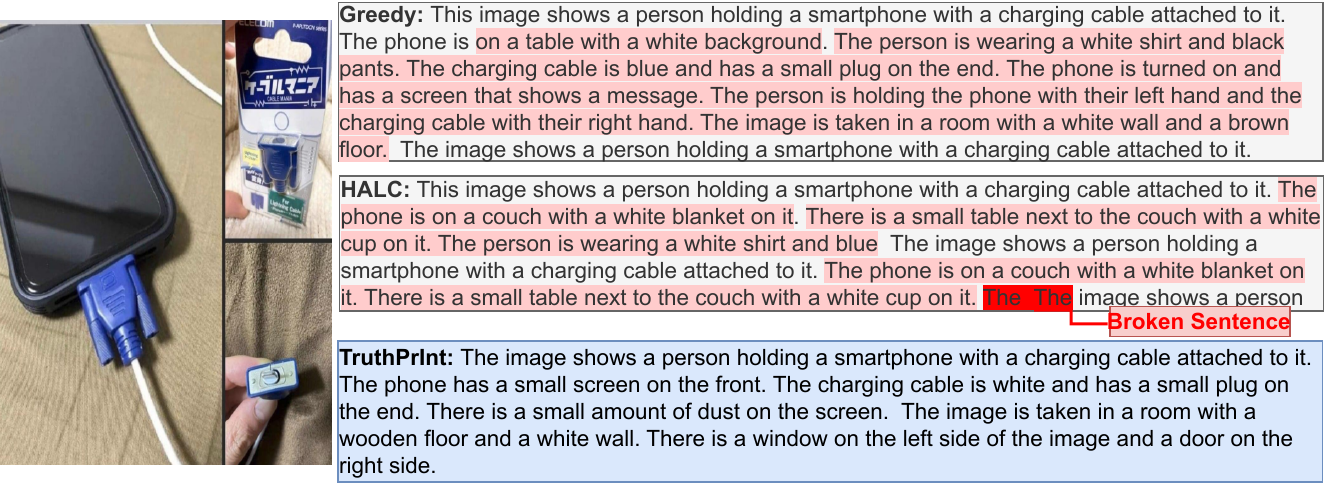}
                \caption{Both Greedy and HALC hallucinate details such as a person wearing a \textbf{white shirt and blue}, along with other nonexistent objects like a ``\textit{phone message}'' and a ``\textit{couch}''. In contrast, \methodname delivers more accurate and truthful descriptions.}
        \end{subfigure}
        
        \begin{subfigure}[b]{\textwidth}
                \includegraphics[width=\linewidth]{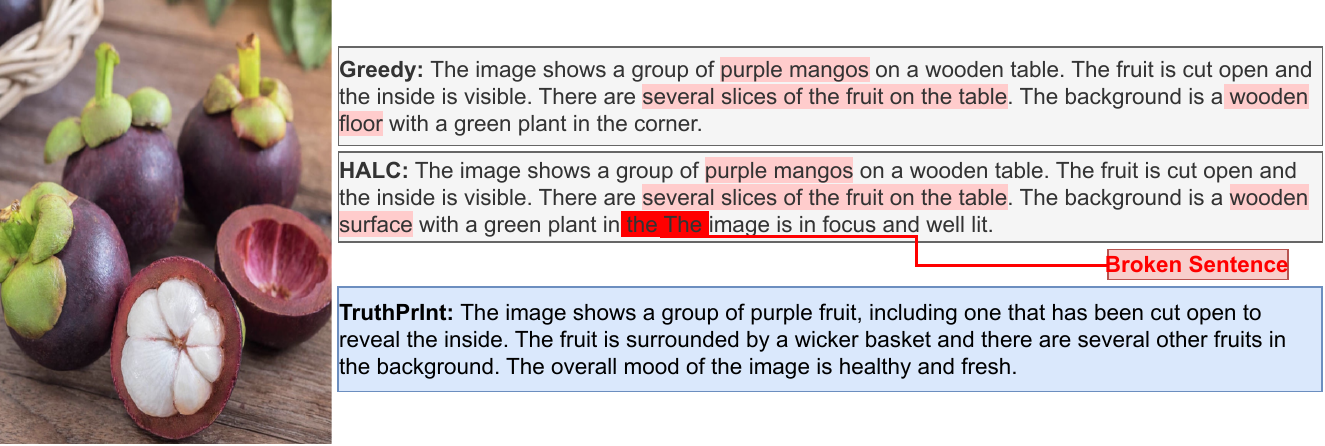}
                \caption{Both Greedy and HALC incorrectly describe the item as a \textbf{purple mango} and further hallucinate details like ``\textit{several slices of this fruit}''. In contrast, \methodname offers a more accurate description, referring to it as \textbf{purple fruits}.}
        \end{subfigure}

        \begin{subfigure}[b]{\textwidth}
                \includegraphics[width=\linewidth]{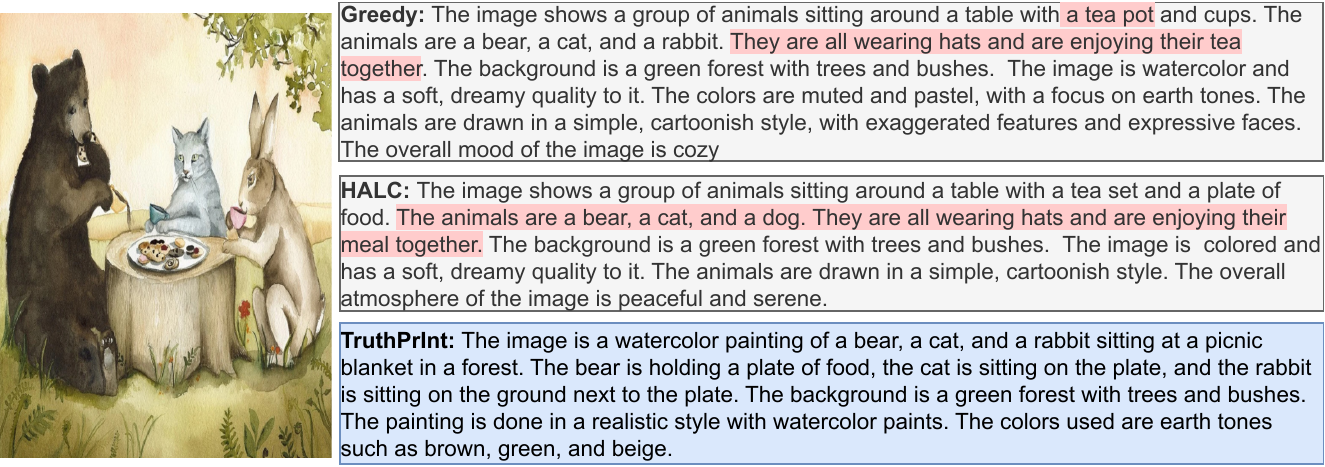}
                \caption{Both Greedy and HALC falsely describe all the animals as \textbf{wearing hats} and provide only limited details about the image. Additionally, \textbf{HALC misidentifies the rabbit as a dog}. In contrast, \methodname delivers accurate descriptions of all the animals and includes additional details such as ``\textit{the bear is holding a plate of food}'' and ``\textit{the colors used are earth tones like brown, green, and beige}''.}
        \end{subfigure}
        \caption{LLaVA-Bench quantitative analysis results.}\label{fig:llavabench}
\end{figure*}

We evaluate our methods and baselines on the LLaVA-Benchmark (In-the-Wild) dataset, manually reviewing the generated responses for these images (~\cref{fig:llavabench}). Our observations reveal that \methodname produces more accurate and truthful descriptions, with greater detail included compared to the baselines.

\end{document}